\documentclass{article}

\usepackage{arxiv}

\usepackage{multirow}
\usepackage[utf8]{inputenc} 
\usepackage[T1]{fontenc}    
\usepackage{hyperref}       
\usepackage{url}            
\usepackage{booktabs}       
\usepackage{amsfonts}       
\usepackage{nicefrac}       
\usepackage{microtype}      
\usepackage{lipsum}		
\usepackage{graphicx}
\usepackage{natbib}
\usepackage{doi}

\title{RN-Net: Reservoir Nodes-Enabled Neuromorphic Vision Sensing Network}

\date{} 					

\author{ \href{https://orcid.org/0009-0003-0780-3367}{\includegraphics[scale=0.06]{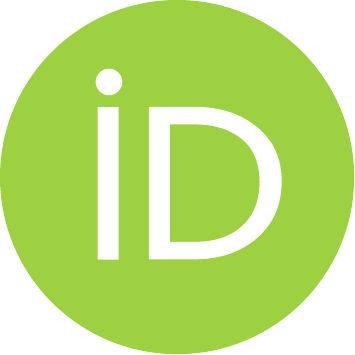}\hspace{1mm}Sangmin Yoo}, \href{https://orcid.org/0009-0000-7027-7476}{\includegraphics[scale=0.06]{orcid.pdf}\hspace{1mm}Eric Yer-Jer Lee}, \href{https://orcid.org/0000-0001-9534-3134}{\includegraphics[scale=0.06]{orcid.pdf}\hspace{1mm}Ziyu Wang}, \href{https://orcid.org/0000-0003-0136-5936}{\includegraphics[scale=0.06]{orcid.pdf}\hspace{1mm}Xinxin Wang}, \href{https://orcid.org/0000-0003-4731-1976}{\includegraphics[scale=0.06]{orcid.pdf}\hspace{1mm}Wei D. Lu}\thanks{Corresponding author}\\
	Department of Electrical Engineering and Computer Science\\
	University of Michigan\\
	Ann Arbor, MI 48105 \\
	\texttt{\{ysangmin, ericylee, ziwa, xinxinw, wluee\}@umich.edu} \\
}

\renewcommand{\headeright}{Technical Report}
\renewcommand{\undertitle}{Technical Report}
\renewcommand{\shorttitle}{RN-Net: Reservoir Nodes-Enabled Neuromorphic Vision Sensing Network}

\hypersetup{
pdftitle={RN-Net: Reservoir Nodes-Enabled Neuromorphic Vision Sensing Network},
pdfsubject={q-bio.NC, q-bio.QM},
pdfauthor={Sangmin Yoo, Eric Yeu-Jer Lee, Ziyu Wang, Xinxin Wang, Wei D. Lu},
pdfkeywords={Neuromorphic Vision, Reservoir Computing},
}

\begin{document}
\maketitle

\begin{abstract}

Neuromorphic computing systems promise high energy efficiency and low latency. In particular, when integrated with neuromorphic sensors, they can be used to produce intelligent systems for a broad range of applications. An event-based camera is such a neuromorphic sensor, inspired by the sparse and asynchronous spike representation of the biological visual system. However, processing the event data requires either using expensive feature descriptors to transform spikes into frames, or using spiking neural networks (SNNs) that are expensive to train. In this work, we propose a neural network architecture, Reservoir Nodes-enabled neuromorphic vision sensing Network (RN-Net), based on dynamic temporal encoding by on-sensor reservoirs and simple deep neural network (DNN) blocks. The reservoir nodes enable efficient temporal processing of asynchronous events by leveraging the native dynamics of the node devices, while the DNN blocks enable spatial feature processing. Combining these blocks in a hierarchical structure, the RN-Net offers efficient processing for both local and global spatiotemporal features. RN-Net executes dynamic vision tasks created by event-based cameras at the highest accuracy reported to date at one order of magnitude lower network size. The use of simple DNN and standard backpropagation-based training rules further reduces implementation and training costs.

\end{abstract}


\section{Introduction}

\label{sec:intro}

Event-based cameras are neuromorphic vision sensors that produce visual signals as asynchronous spikes.\cite{Gallego2022} An event camera produces a spike when and only when a momentary pixel intensity difference exceeds a threshold, which can offer better energy efficiency and latency when compared with conventional cameras that produce data at a constant frame rate even when the scene is stationary during video recording.\par

Various visual tasks have been implemented using event-based cameras. Earlier datasets were generated by reproducing conventional image recognition tasks such as MNIST\cite{Orchard2015}, CIFAR-10\cite{Li2017-CIFARDVS}, and Caltech 101\cite{Orchard2015}, where stationary images were placed in front of the camera and moved around to create pixel intensity differences along time. Objects in the tasks are dynamic in their locations with their shape fixed over time, which requires networks to be capable of classifying objects regardless of their positional change. Beyond the positional dynamic movement, behaviorally more dynamic datasets, such as DVS128 Gesture\cite{Amir2017}, N-CARS\cite{Sironi2018}, DVS Lip\cite{Tan2022} were later created by the event camera to fully utilize the camera’s strength where the temporal evolutions of both the object’s shape and movement are critical.\par

\begin{figure*}[!t]
  \includegraphics[width=\linewidth]{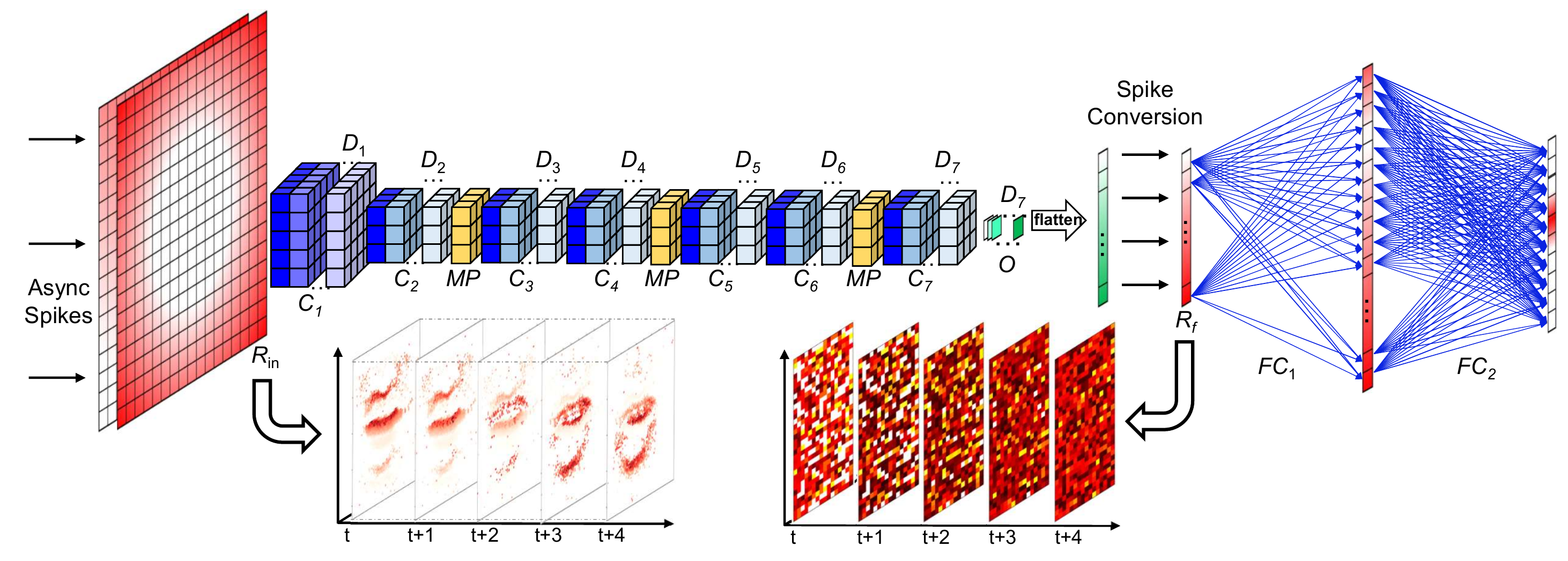}
  \caption{RN-Net structure. $\emph{R}_{in}$ and $\emph{R}_{f}$ are reservoir layers for local and global temporal feature encoding, respectively. Bottom left: outputs from $\emph{R}_{in}$ from a representative input in the DVS Lip dataset. Bottom right: outputs from $\emph{R}_{f}$. Outputs from $\emph{R}_{f}$ are reshaped in 2D for better visualization. Deeper color in $\emph{R}_{in}$, $\emph{R}_{f}$ and output layers of Convolution (Conv) blocks and Fully-Connected ($\emph{FC}_{1-2}$) layers represents a higher analog value. Hidden layers within Conv blocks are not presented. The DNN structure for DVS Lip dataset is representatively illustrated. $\emph{C}_{N}$, $\emph{D}_{N}$, $\emph{MP}$, $\emph{FC}_{N}$ represent N-th Conv layer, kernel Depth of N-th Conv layer, Max Pooling layer and N-th FC layer, respectively.}
  \label{fig:RN-Net Structure}
  
\end{figure*}

To process either the positionally dynamic datasets or the behaviorally and morphologically more dynamic datasets, many efforts have been made based on deep neural networks (DNNs) consisting of convolution and fully connected layers or spiking neural networks (SNNs).\cite{Zhu2022,Deng2022,Yao2021,Fang2021-deep,Fang2021-incorporating,Zheng2021,Wu2022-LIAF,kaiser2020,kugele2020,Shrestha2018,Lagorce2017,Sironi2018,Amir2017,Tan2022,Xiao2020,Feng2020,Martinez2020,Wang2019,Wang2019-EV-gait,Gehrig2019,Wang2022-event-stream} However, DNNs require additional recurrent units which increase memory and training costs to process temporal information embedded in the datasets due to the lack of temporal processing capability and diminish the merits of asynchronous operation of the event-based camera due to its synchronized processing. SNNs allow temporal processing, however the necessity of BPTT for proper training increases the training cost.\cite{werbos1990} As the solution, feature descriptors such as Time Surface (TS) were proposed.\cite{Lagorce2017,Sironi2018,Grimaldi2022} It is cost-effective and powerful in that it encodes only the latest temporal information and can be integrated with DNN structures, enabling gradient-based training. However, its latest information-dominant encoding fashion discards useful information prior to the last spike, which may be essential for features that evolve over a longer history.

In this paper, we introduce a neural network architecture, Reservoir Nodes-enabled neuromorphic vision sensing Network (RN-Net), that leverages simple reservoir layers and DNN blocks for temporal and spatial processing, respectively. Two reservoir layers are employed, with the one at the front encoding the temporally local information on the sensor by directly receiving asynchronous events (spikes) and the one at the back encoding the temporally global information without the expensive recurrent units, respectively, both leveraging native dynamics of reservoir nodes (RNs). The DNN blocks find spatial information in data temporally processed by the reservoirs, resulting in spatiotemporal output in the end.\par

Specifically, the nature of short-term memory (STM) memristors implementing RNs allows RNs to work as native feature descriptions of temporal information in all prior spikes produced by the event camera or hidden layers, without dedicated memory and CMOS logic circuits usually required by recurrent units or complex feature descriptor algorithms.\cite{Du2017,Moon2019,Moon2021,Chang2011} RNs in the network can be thought of as analogous to cells in a retina which directly sense and encode raw and asynchronous visual inputs, and transmit them to the brain.\cite{Masland2012} Besides, the better encoding capability of RNs beyond the latest spike than TS that focuses only on the latest one enables RN-Net to perform better with a much smaller network size than others. (Figure 1) An example of the proposed RN-Net is shown in Figure~\ref{fig:RN-Net Structure}.\par
The main contributions of this work can be summarized below:\par
\begin{itemize}
\item We implement a neural network with multiple reservoirs and DNN blocks that process temporal and spatial information embedded in asynchronous event streams generated by event-based cameras, respectively.
\item On-sensor RNs based on STM memristors offer richer temporal spike encoding at a lower cost, which makes DNN blocks simpler, leading to more efficient operation and training. 
\item RN-Net performs CIFAR10-DVS, N-Caltech 101, DVS128 Gesture, and N-CARS tasks at the highest accuracy reported to date and DVS Lip task at one of the highest accuracy at one order of magnitude lower network size than other networks with similar capacity.
\end{itemize}

\section{Backgrounds}

\label{sec:Backgrounds}
\subsection{Event-Based Dataset}
\label{sec:Event-Based Dataset}

In the early years, conventional static vision tasks were reproduced by the event-based camera by moving images in front of the camera to make changes in pixel intensity.\cite{Orchard2015,Li2017-CIFARDVS} Beyond the conversion, dedicated datasets have been created to maximize the strengths of event-based cameras.\cite{Amir2017,Tan2022,Bi2019-graph,Mueggler2016, Gehrig2021,Serrano-Gotarredona2015} Among them, we use CIFAR10-DVS, N-Caltech 101, DVS128 Gesture, N-CARS, and DVS Lip datasets in this work. The CIFAR10-DVS and N-Caltech 101 are object detection tasks, The DVS128 Gesture dataset is for human action recognition, the N-CARS is for car classification in a real-world setting, and the DVS Lip dataset is for word recognition based on lip motions of speaking participants. Data of the first two datasets are dynamic in object's location over time, while the rest of the datasets include temporally dynamic data in both shape and movement. The details of the datasets are summarized in Table~\ref{tab:Details of datasets}. We split datasets that are not origianlly divided into train/test data 9:1.\par

\begin{table}[h]
 \centering
 \caption{Details of datasets used for RN-Net demonstration.}
  \begin{tabular}{lccl}
    \hline
    Dataset        & Train & Test & Category \\
    \hline
    CIFAR10-DVS    & 9000  & 1000 & 10 (Objects)\\
    N-Caltech 101  & 7839  & 870  & 101 (Objects)\\
    DVS128 Gesture & 1077  & 264  & 11 (Actions)\\
    N-CARS         & 15422 & 8607 & 2  (Car/Background)\\
    DVS Lip        & 14896 & 4975 & 100 (Words)\\
    \hline
  \end{tabular}
  \label{tab:Details of datasets}
\end{table}

\subsection{Related Works}

There have been various attempts to better execute the event-based tasks, utilizing DNNs consisting of convolution and fully connected layers, and SNNs have been demonstrated.\cite{Zhu2022,Deng2022,Yao2021,Fang2021-deep,Fang2021-incorporating,Zheng2021,Wu2022-LIAF,kaiser2020,kugele2020,Shrestha2018,Lagorce2017,Sironi2018,Amir2017,Tan2022,Xiao2020,Feng2020,Martinez2020,Wang2019,Wang2019-EV-gait,Gehrig2019,Wang2022-event-stream}. Given that DNNs are specialized in processing static spatial data, recurrent units, such as long short-term memory (LSTM)\cite{Hochreiter1997,Graves2013}, gated recurrent unit (GRU)\cite{Chung2014} and bidirectional gated recurrent unit (BiGRU)\cite{Tan2022,Xiao2020,Feng2020,Martinez2020} are typically required to process temporal information hidden in the sequence (global) of momentary (local) features. The momentary features are processed synchronously in temporally local frames which are created either by simple accumulation of spikes within a prefixed time range\cite{Wang2019-EV-gait} or using input representation algorithms like Graph construction\cite{Wang2022-event-stream,Bi2019}, 3D point cloud\cite{Wang2019}, Event frame\cite{Rebecq2017},  Event spike tensor\cite{Gehrig2019} and voxel grid.\cite{Tan2022,Zhu2019}. Synchronized processing requires storing and analyzing a large number of events as a pre-processing step, which diminishes the advantages of asynchronous and sparse spike generation features of event-based cameras. Recurrent units require storing multiple state data for each node, causing increased training costs.\par

SNNs store temporal information in the neuron dynamics using models such as leaky integrate-and-fire (LIF) neurons\cite{Yao2021,Wu2022-LIAF} and can be trained using gradient-based approaches such as backpropagation through time (BPTT).\cite{werbos1990} To address the non-differentiability of spikes, surrogate approaches that replace spikes with neuron membrane potential have been developed.\cite{Neftci2019} However, BPTT requires backpropagation through both the network layers and time, which technically makes the network n times larger when unfolded in time, where n is the number of timesteps going back in time. As a result, BPTT is expensive to train. Other works have employed bio-inspired local learning rules, such as spiking timing dependent plasticity (STDP), for less training cost.\cite{STDP-trained-SNN} Although these methods result in significantly improved efficiency, the local learning rule is generally worse than BPTT in training quality.\par

To use spikes directly, feature descriptors such as TS\cite{Lagorce2017,Sironi2018,Grimaldi2022} were proposed. TS stores only the last spike event for every pixel and converts the time to the last spike information into an analog value by resetting the amplitude of each node to 1 every time it receives a spike and then relaxes following a decay function:
\begin{equation}
S(t)=e^{-\frac{t-T(t)}{\tau}}
\end{equation}
where \emph{S(t)} is an analog vector representing a node on the time surface, \emph{t} is the current time, \emph{T(t)} is the time information of the last spike received by the node, and $\tau$ is a pre-defined time constant.\cite{Grimaldi2022,tapiador-morales2020} The TS conversion allows the encoded data in the analog surface to be processed with DNN and trained using gradient-based training while reducing memory and computing costs compared to other feature descriptors, since only the last spike instant needs to be recorded. However, since TS only stores the last spike, it cannot handle spatiotemporal features whose correlation is beyond its temporal neighbors, which is common in real-world problems. Additionally, storing the last spike timing information for each pixel (node), and calculating the analog state based on equation (1) still incurs substantial costs. More advanced approaches such as Leaky Surface were subsequently developed to encode the temporal information beyond the last spike\cite{Cannici2019}. However, expensive pre-processing is still required, for example, to store the previous node state and time elapsed from the latest spike, and to calculate the new state for every pixel at every time instant.\par

\subsection{Reservoir Nodes}

We note that in a reservoir computing (RC) system, the reservoir maintains short-term memory and performs nonlinear transformation (encoding) of the temporal input data into the reservoir states, represented by the states of the reservoir nodes (RNs). RC systems have been efficiently implemented in hardware using devices such as memristors for vision (MNIST handwritten digits recognition), speech (NIST TI46) and Time-series forecasting (Mackey-Glass time series) tasks.\cite{Du2017,Moon2019} In these systems, the reservoir nodes encode the spikes’ spatiotemporal information naturally following the internal device dynamics, without any external memory or arithmetic and logic units (ALUs). By leveraging the internal device and circuit dynamics to process temporal data, these implementations have shown excellent energy efficiency and performance.\cite{Du2017,Moon2019,Li2017,Farronato2022,Fang2021}\par
Generally, due to the STM property, RNs will be affected more strongly by the near history events and weakly by far history events, with the extent of the non-linearity determined by the internal RN time constant.\cite{Chang2011,Du2015} Inspired by this principle, we hypothesize that RNs can be directly used to encode temporal spike data similar to what TS aims to accomplish, but at a lower cost and better encoding capability beyond the last spike since RNs \emph{non-linearly} accumulate all prior incoming spike information.

\section{Methods}
\label{sec:Methods}

\subsection{Event Encoding with Reservoir Nodes}

A reservoir node can be implemented using only a single STM memristor, making hardware implementation very light weight.\cite{Du2017,Moon2019} In a STM memristor, the node state (i.e. device conductance) is natively excited by the incoming spikes and relaxes in between the spikes,\cite{Chang2011,Du2015} as described by the following equation:
\begin{equation}
\label{equ:dynamics}
G_{t}=P_{c}*(G_{max}-G_{t-1})*\delta_{spk}(t)+G_{t-1}*e^{-\frac{1}{\tau}}
\end{equation}
where $\emph{G}_{t}$ is the node state at time $\emph{t}$, $\emph{P}_{c}$ is a potentiation factor, $\emph{G}_{max}$ is the upper bound of the node state, $\tau$ is the characteristic relaxation time constant, and $\delta_{spk}$\emph{(t)} is a delta function representing a spiking event:
\begin{equation}
\label{equ:spiking delta}
  \delta_{spk}(t)=\left\{
    \begin{array}{ll}
      0, & \mbox{when no spike}.\\
      1, & \mbox{when receiving spike}.
    \end{array}
  \right.
\end{equation}
$\delta_{spk}$\emph{(t)} can represent incoming events from an event-based camera or spikes from preceding layers within the network.\par

\begin{figure}[h]
  \centering
  \includegraphics[scale=0.4]{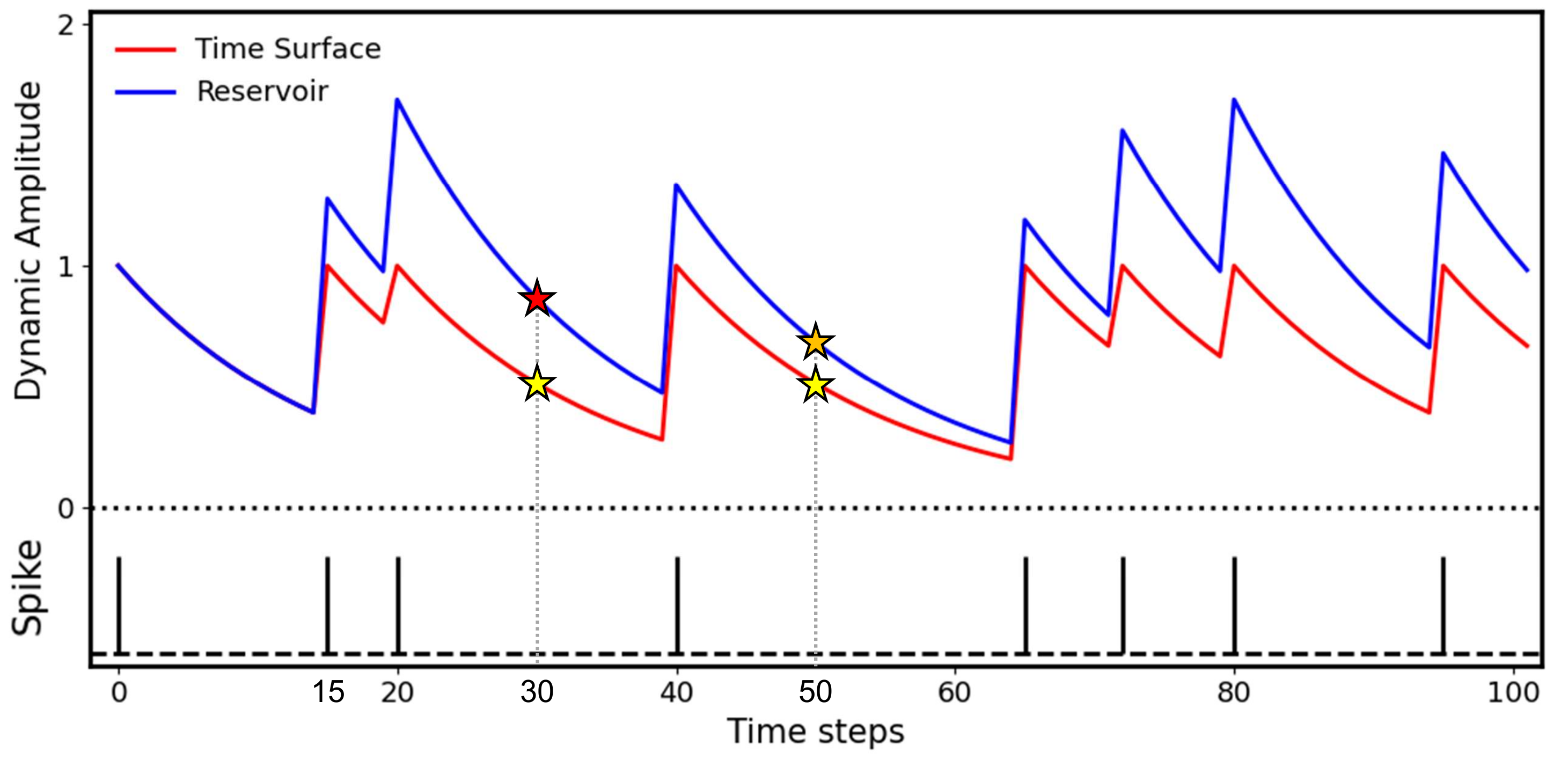}
  \caption{Dynamics of a reservoir node (blue) and a time surface node (red) under identical spikes (black) shown below.}
  \label{fig:Dynamics}
\end{figure}

Figure~\ref{fig:Dynamics} shows the comparison of the RN approach implemented with a single STM memristor based on equation (2) versus the TS approach introduced in Section 2.2. Both approaches convert the spiking patterns into an analog state that can be processed by subsequent DNN blocks. Compared to TS encoding which resets the state to 1 after each spike, the RN implementation allows longer-term history to be represented in the state in a non-linear fashion. For example, at $\emph{t}$=50, the TS output is identical to that at $\emph{t}$=30, since both values were reset to 1 10 time steps earlier (at $\emph{t}$=20 and 40, respectively). However, this representation missed the differences in the two temporal sequences (e.g. an additional preceding spike at $\emph{t}$=15). On the other hand, the RN implementation clearly differentiates the two cases as all inputs before the current time are accumulated non-linearly. Combined with the hardware efficiency, e.g., equation (2) is natively implemented in a single device owing to internal device physics (ionic and electronic dynamic processes)\cite{Du2017,Moon2019} without the need of additional dedicated memory (to store $\emph{t}$ and \emph{T(t)}) and ALUs (to execute equation (1)), we believe the RN-Net to be a suitable solution for asynchronous, real-time processing of event data.\par

The proposed RN-Net operates by directly taking asynchronous spikes as they are generated. The memristor-based RNs located in the first part of RN-Net, each of which is connected to a pixel, autonomously transform the temporal spikes into analog values following Equation~\ref{equ:dynamics} on the sensor in real-time, analogous to visual inputs encoded by cells in a retina.\cite{Masland2012} The states of the RNs are retrieved only when the network operates a forward pass, as shown in Figure~\ref{fig:RN-Net Structure}. In the process, the temporal events from a pixel output are encoded as the state of the corresponding RN node in the $\emph{R}_{in}$ layer, and the spatial information is preserved in the $\emph{R}_{in}$ layer since each RN node is independently connected to a corresponding pixel. As a result, the states of all nodes in the $\emph{R}_{in}$ layer encode the spatiotemporal information of the inputs. The values of the $\emph{R}_{in}$ node states can then be read out and processed using conventional DNN blocks. Notably, owing to the discrete memristor dedicated to a pixel, RN-Net is free of parasitic resistance-capacitance effects of the crossbar format that is common for memristor applications. In the second half of RN-Net, after the convolution blocks and the spikes conversion layer, the spatiotemporal features are again embedded in the spikes. Instead of using the number of spikes to perform classification in the subsequent FC layers, we chose to use another RN layer ($\emph{R}_{f}$ layer in Figure~\ref{fig:RN-Net Structure}) to encode the spatiotemporal features discovered by the first half of the network. At the end of a video input clip, the states of $\emph{R}_{f}$ capture the long-term temporal information and are supplied to the FC layers for classification.\par

\begin{figure}[h]
  \includegraphics[width=\linewidth]{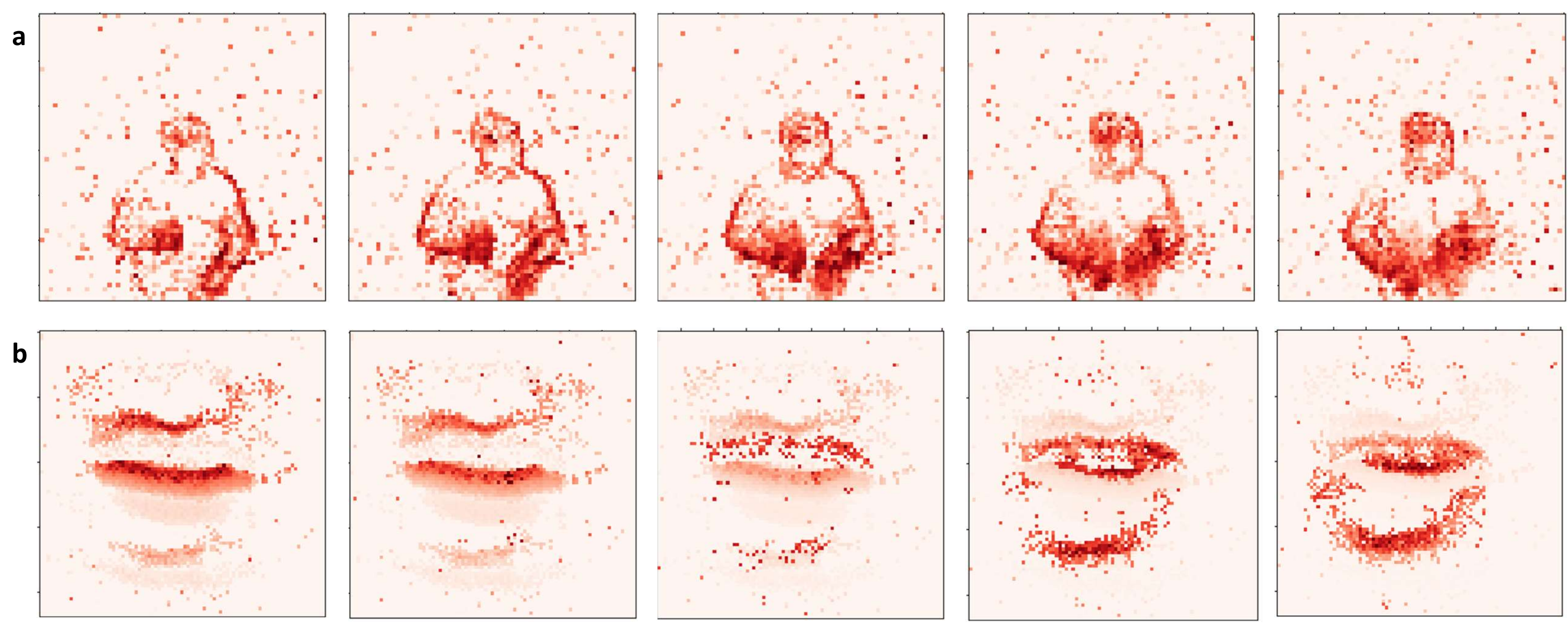}
  \caption{Temporally consecutive states of the input reservoir nodes, responding to asynchronous events in the (a) DVS128 Gesture and (b) DVS Lip datasets. Each state is retrieved at a constant time interval of 30ms. Deeper red represents a higher amplitude value of an RN state.}
  \label{fig:Sample RN states}
\end{figure}

Figure~\ref{fig:Sample RN states} shows examples of 5 temporally consecutive states of the input layer ($\emph{R}_{in}$) RNs, responding to asynchronous events from two representative datasets. The states are obtained at a constant time interval (i.e. 30ms). Frequent input spikes lead to stronger RN states due to the excitation term (the first term of Equation~\ref{equ:dynamics}), while RN states natively relax for inputs that are temporally too far from the current time instant due to the internal decay term (the second term of Equation~\ref{equ:dynamics}). These properties allow RNs to capture temporal features of inputs independently and the $\emph{R}_{in}$ layer possesses spatial features when the RNs are gathered, as shown in Figure~\ref{fig:Sample RN states}. For example, in Figure~\ref{fig:Sample RN states}b, both the speaking motion and movement (top to bottom) of the lips can be captured by the RNs’ states in the $\emph{R}_{in}$ layer. Temporally fast movements are reflected in a single moment (e.g. the 3rd plot in Figure~\ref{fig:Sample RN states}b), while temporally slow movements are reflected in multiple moments captured at the different time instants (e.g. the first two plots).\par
\subsection{Model Architecture}

The RN-Net is formed sequentially by an input RN layer ($\emph{R}_{in}$) as a feature descriptor for temporally local encoding, convolution (Conv) layers ($\emph{C}_{N}$) for spatial processing, a spike conversion (SC) layer, another RN layer ($\emph{R}_{f}$) for temporally global encoding, and multi-layer perceptron (MLP) ($\emph{F}_{N}$) for classification. Figure~\ref{fig:RN-Net Structure} illustrates the overall architecture of the proposed network for the DVS Lip dataset.\par
The $\emph{R}_{in}$ layer has as many RNs as the output dimension of the event-based camera. For example of the DVS128 Gesture dataset with 2x128x128 pixels, where 2 represents the polarity of the events, 32768 (2x128x128) independent nodes are used to form the $\emph{R}_{in}$ layer.
Each RN in the $\emph{R}_{in}$ layer processes input spikes asynchronously from a corresponding pixel in the event camera in real time without any preprocessing, following Equation~\ref{equ:dynamics}. Examples of the outputs generated by $\emph{R}_{in}$ are shown in Figure~\ref{fig:Sample RN states}.\par
Depending on the application, different temporal resolutions can be chosen by adjusting $\emph{P}_{c}$ and $\tau$ of Equation~\ref{equ:dynamics} and the interval of $\emph{R}_{in}$ state acquisitions. A shorter time constant $\tau$ and more frequent acquisitions (shorter interval) produce temporally finer outputs, while a longer time constant and less frequent acquisitions (longer interval) produce spatially more detailed outputs owing to more input spikes and the slower internal decay.\cite{Tan2022} Accordingly, a shorter acquisition interval results in more frequent activations for the following layers. The convolution layers ($\emph{C}_{1}$-$\emph{C}_{7}$ with depths $\emph{D}_{1}$-$\emph{D}_{7}$ in Figure~\ref{fig:RN-Net Structure}) then process the encoded spatiotemporal features in $\emph{R}_{in}$ at a constant time interval (i.e. 30ms). Similar to Conv blocks in conventional DNNs, the output $\emph{O}$ from the Conv layers reflects the spatial features existing in the $\emph{R}_{in}$ states, and in this case, the spatiotemporal features embedded in the spike stream within the time interval.\par

\begin{figure}[h]
  \includegraphics[width=\linewidth]{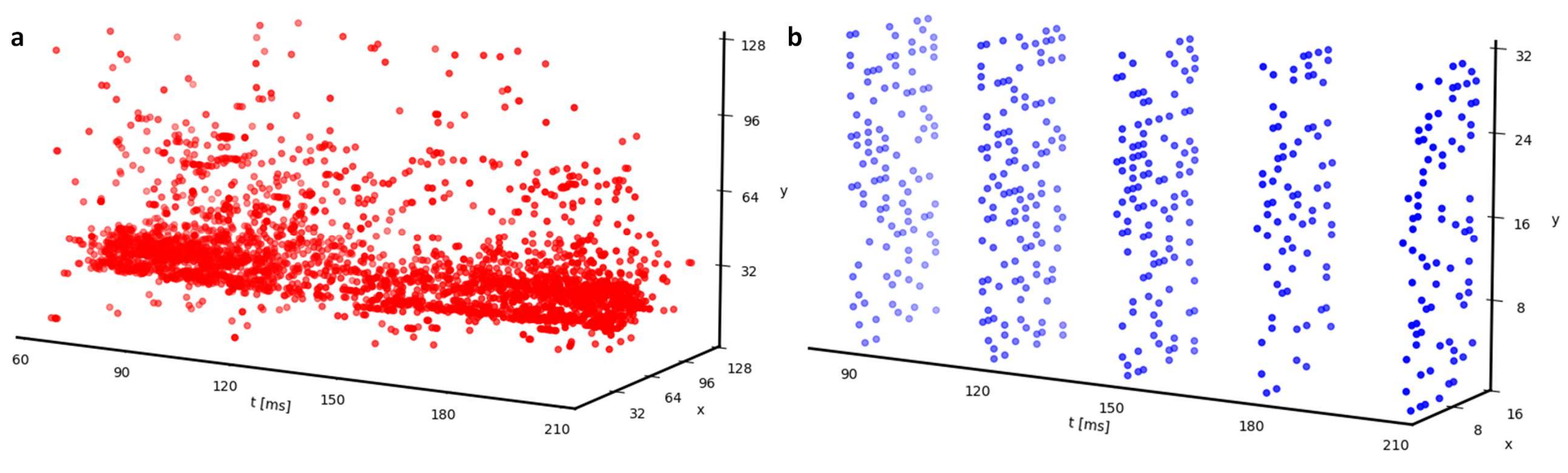}
  \caption{(a) Visualization of asynchronous input spikes and (b) spikes generated after the spike conversion (SC) layer.}
  \label{fig:visualization of SC}
\end{figure}

Output $\emph{O}$ is then flattened and sent to the spike conversion SC layer. We use a predetermined threshold to convert the outputs $\emph{O}$ from the Conv layers into spikes in the SC layer. Spikes generated after the SC layer are shown in Figure~\ref{fig:visualization of SC}. Thus, the spikes after SC represent the local spatiotemporal features captured by the $\emph{R}_{in}$ layer and the Conv layers, and become much sparser compared with the original spiking inputs.\par

The spikes from the SC layer are then supplied to the RNs in the $\emph{R}_{f}$ layer for global spatiotemporal feature encoding and subsequent classification in MLP. Similar to the RNs in the $\emph{R}_{in}$ layer, RNs (e.g. also implemented with STM memristors) in $\emph{R}_{f}$ naturally potentiates/relaxes with the presence/absence of spikes from the SC layer. The state of $\emph{R}_{f}$ thus represents the historical (global) spatiotemporal features by encoding the processed local spatiotemperal features over time, and is then used by the subsequent fully-connected (FC) layers of MLP ($\emph{FC}_{1}$-$\emph{FC}_{2}$ in Figure~\ref{fig:RN-Net Structure}) to perform final classification functions. Similar to the RNs in the $\emph{R}_{in}$ layer, RN states in $\emph{R}_{f}$ are retrieved at a certain time interval (i.e. 0.3s, which is longer than that used in $\emph{R}_{in}$) over the whole video clip and fed to the FC layers.\cite{Alomar2015} The final decision is made based on accumulated potentials through multiple FC feedforwards. (i.e. 5)\par

\subsection{Training Method}

We train RN-Net with standard backpropagation, using the output potentials calculated by $\emph{FC}_{2}$ at the end of data presentation. Unlike BPTT that calculates error and gradient across each timestep\cite{werbos1990}, error and gradient of RN-Net is calculated only once for one input training video, leading to lower training cost. To address the non-differentiability arising from spike generation in the SC layer, we adopted surrogate gradient for the SC layer. Since the surrogate is used only in one layer of the network, we expect the error introduced by this approach to be lower when compared with SNN approaches which require surrogates at all layers.\par
We also applied several data augmentation techniques, such as random/center crop, horizontal flip and Gaussian noise to minimize overfitting effects. In the case of random/center crop, we referred to existing works.\cite{Tan2022,Feng2020} During training, we center-cropped the original input dimension (128 x 128) to 96 x 96 and then random cropped them to 76 x 76, while the inputs are directly center-cropped to 76 x 76 in the testing phase. Horizontal flip with a probability of 0.5 and Gaussian noise with standard deviation of 5$e^{-4}$ were used during training for the same purpose. These techniques helped generalize the input data and reduce overfitting during training. These techniques were used only during training of RN-Net for the DVS Lip.\par

\section{Experimental Section}
\label{sec:Experiments}
\subsection{Experiment Setup}

The DVS128 Gesture dataset consists of repetitions of the same motion of a participant along the clip, while lip motions in the DVS Lip dataset are not repetitive, as shown in Figure~\ref{fig:Sample RN states}. As a result, a fraction of a clip should be sufficient to classify the action in DVS128 Gesture, while the whole clip is necessary for DVS Lip classification. Based on this understanding, we only used the first 1.5s for all videos in the DVS128 Gesture dataset during training and inference, corresponding to ~10\% of the longest video (15.5s). For CIFAR10-DVS, N-Caltech 101, and N-CARS datasets, the first 0.6s, 0.3s, and 90ms are used, respectively when the videos are longer than the lengths for the same purpose. The DVS Lip dataset has variable input lengths due to the irregular length of the words and the unique speaker’s traits, with most video clips (99\%) having lengths between 0.75s and 1.5s. To make the data size regular across the dataset, we chose to keep the captured $\emph{R}_{in}$ outputs at 50 by simply applying null (no spike) for clips shorter than 1.5s and clipping videos longer than 1.5s during DVS Lip training and inference.\par

For all the datasets, the same potentiation factor $\emph{P}_{c}$ (0.5) and time constant $\tau$ (60ms) in Equation~\ref{equ:dynamics} are used for the RNs in $\emph{R}_{in}$. The values are referred to prior works.\cite{Du2017,Moon2019} The states of $\emph{R}_{in}$ are captured every 30ms and sent to the subsequent Conv layers. We note that we used one set of state retrieval frequency and time constant chosen for all the datasets to demonstrate the potential of RN-Net, although tuning the time constants and feeding data at different intervals for a specific task can further enhance the performance.\cite{Du2017}\par

Due to the different input dimensions and the different number of categories in the datasets, the convolution layers and FC layers in RN-Net are configured accordingly. The detailed network configurations for the datasets are summarized in Table~\ref{tab:network configuration}.\par

\begin{table}[h]
 \caption{Network configuration of RN-Net. The left half presents the architecture for DVS Lip and the right half presents the architecture for the rest. The last column of each half represents the output dimension of the network for datasets other than N-Caltech 101. The 6th convolution layer and the following max-pooling layer in the right half are exclusively applied for N-Caltech 101 due to its larger input dimension (240x180) than the others (128x128).}
  \begin{tabular}{ccccc|ccccc}
    \hline
    \multicolumn{5}{c}{DVS Lip} & \multicolumn{5}{c}{Others}\\
    \hline
    Layer     & Kernel & Out     & Pad/   & Output & Layer     & Kernel & Out     & Pad/   & Output\\
              & Size   & channel & Stride &   Dim &           & Size   & channel & Stride &   Dim\\
    \hline
    $\emph{R}_{in}$ & - & 2 & - & 76 x 76 & $\emph{R}_{in}$ & -  & 2  & - & 128 x 128\\
    Conv1   & 5 & 64  & 0 / 2 & 36 x 36 & MaxPool  & 2 &  2  & 0 / 2 & 64 x 64\\
    Conv2   & 3 & 128 & 1 / 1 & 36 x 36 & Conv1    & 3 &  64 & 0 / 1 & 62 x 62\\
    MaxPool & 3 & 128 & 0 / 2 & 17 x 17 & MaxPool  & 3 &  64 & 0 / 2 & 30 x 30\\
    Conv3   & 3 & 128 & 1 / 1 & 17 x 17 & Conv2    & 3 & 128 & 1 / 1 & 30 x 30\\
    Conv4   & 3 & 256 & 1 / 1 & 17 x 17 & MaxPool  & 3 & 128 & 0 / 2 & 14 x 14\\
    MaxPool & 3 & 256 & 0 / 2 & 8 x 8   & Conv3    & 3 & 256 & 0 / 1 & 12 x 12\\
    Conv5   & 3 & 256 & 1 / 1 & 8 x 8   & Conv4    & 3 & 512 & 1 / 1 & 12 x 12\\
    Conv6   & 3 & 512 & 1 / 1 & 8 x 8   & MaxPool  & 3 & 512 & 0 / 2 & 5 x 5 \\
    MaxPool & 3 & 512 & 0 / 2 & 3 x 3   & Conv5    & 3 & 512 & 0 / 1 & 3 x 3 \\
    Conv7   & 3 & 512 & 0 / 1 & 1 x 1   & MaxPool  & 3 & 512 & 0 / 1 & 1 x 1 \\
    $\emph{R}_{f}$ & - & 512 & - & -    & Conv6    & 3 & 512 & 0 / 1 &  -    \\
    &&&&& MaxPool  & 3 & 512 & 0 / 1 &  -    \\
    \hline
    FC1   & - & 512 & -     & - & FC1      & - & 512 & -     & -      \\
    FC2   & - & 100 & -     & - & FC2      & - & 11 & -     & -       \\
    \hline
  \end{tabular}
  \label{tab:network configuration}
\end{table}

The output of a DNN block consists of a convolution and a pooling layer (if exists), following:
\begin{equation}
O_{N}=f(BN(MP(C(I_{N}))))
\end{equation}
where $\emph{O}_{N}$ is the output of N-th layer, $\emph{f(x)}$ is a ReLU activation function, $\emph{BN(x)}$ is a batch normalization function, $\emph{MP(x)}$/$\emph{C(x)}$ is a maxpooling/convolution operation, and $\emph{I}_{N}$ is input of the N-th layer.\cite{Ioffe2015}\par
After the last convolution layer, the outputs are converted to spikes for subsequent temporal feature encoding at $\emph{R}_{f}$.\cite{Eshraghian2022-training} The SC layer generates spikes by comparing the analogue values in $\emph{O}$ (Figure~\ref{fig:RN-Net Structure}) with a pre-fixed threshold value, set as 0.3.\par

\begin{figure}[h]
  \centering
  \includegraphics[width=0.9\linewidth]{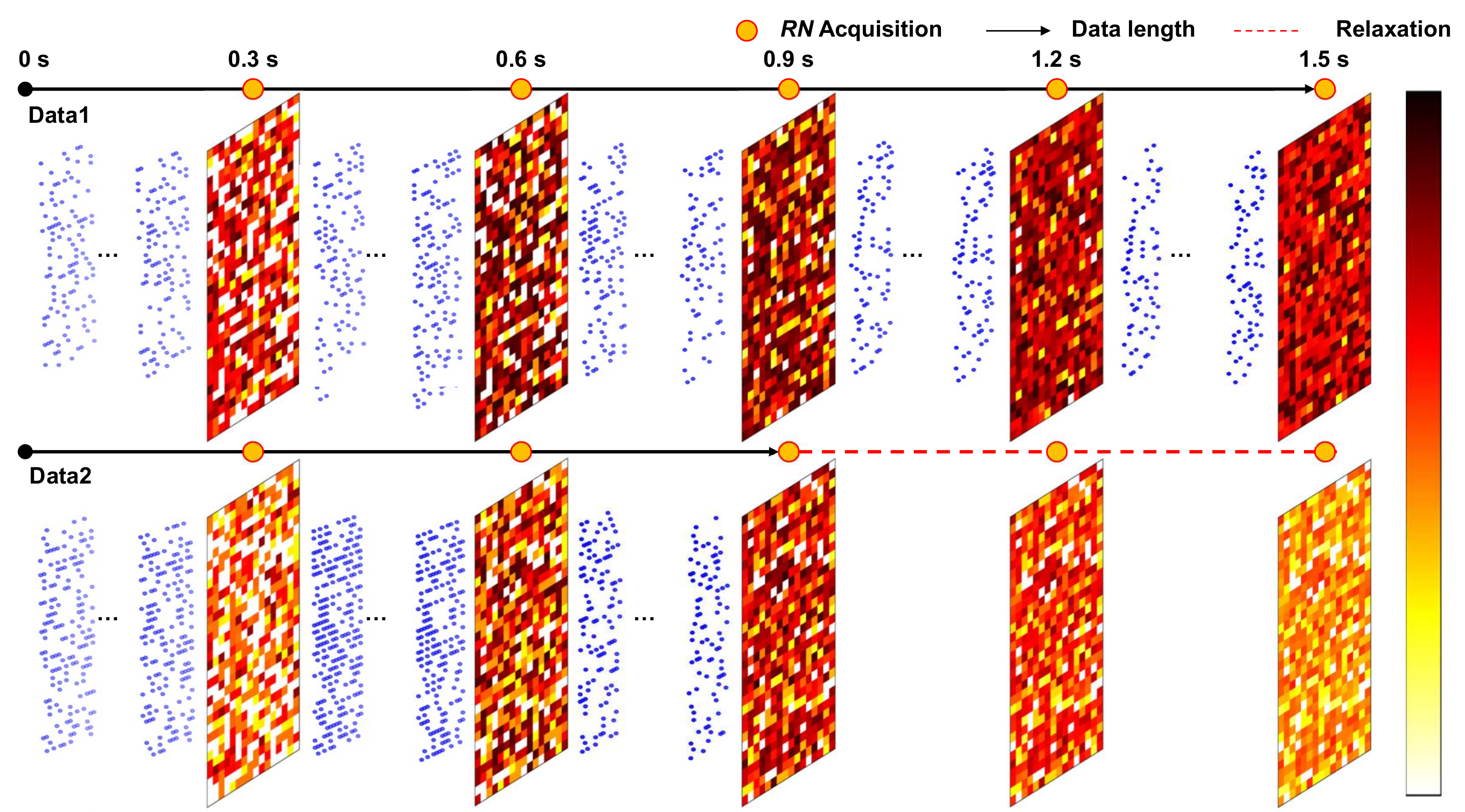}
  \caption{Visualization of outputs from $\emph{R}_{f}$ over the whole 1.5s clip, along with spikes generated from the spike conversion layer. For Data2 whose input video length is only 0.9s, the RN states will continue to relax and still used for classification.}
  \label{fig:Rf outputs}
\end{figure}

RNs in the $\emph{R}_{f}$ layer encode the global spatiotemporal features. $\emph{P}_{c}$ and $\tau$ for the RNs in $\emph{R}_{f}$ are set as 0.3, 2s and 0.1, 2s for DVS Lip and other datasets, respectively. A longer $\tau$ is used in $\emph{R}_{f}$ than in $\emph{R}_{in}$ to process the longer temporal correlations. The RN states in the $\emph{R}_{f}$ layer are retrieved every 0.3s for both DVS128 Gesture and DVS Lip, every 0.12s for CIFAR10-DVS, every 60ms for N-Caltech 101, and every 30ms for N-CARS (corresponding to 5 $\emph{R}_{f}$ acquisitions for all other than N-CARS (3)). In the case of clips shorter than a fixed video length, we let RNs in $\emph{R}_{f}$ continue relaxing after the last meaningful input without padding the input data with new spikes (e.g., through repeating the clip). These relaxed states are still captured following the normal schedule and fed to the classification layers. Examples of RN states at different time instants in DVS Lip task are shown in Figure~\ref{fig:Rf outputs}, along with the spikes they receive from the SC layer. For better visualization, the data are reshaped in 2D format. As shown in the figure, even without any new inputs (e.g., after 0.9s for Data2), the RN states do not immediately decay to zero due to the slow decay term in Equation~\ref{equ:dynamics}, and the evolution of the RNs still represents useful information. This approach also simplifies training and inference processes and allows the system to handle inputs with irregular lengths.\par

MLP consists of 2 FC layers, an activation function, and a batch normalization layer, following: 
\begin{equation}
O_{F}=FC(f(BN(FC(I))))
\end{equation}
where $\emph{O}_{F}$ is the output of the block, $\emph{f(x)}$ is a ReLU activation function, $\emph{BN(x)}$ is a batch normalization function, $\emph{FC(x)}$ is an FC layer, and $\emph{I}$ is the input to the FC layer. In RN-Net, the output of MLP is used as the final output.\par

The Pytorch framework\cite{NEURIPS2019_bdbca288} was used for all the experiments with methods described above. SoftMax function was chosen to calculate the probability of an output neuron in the final output, and the neuron with the largest probability was selected as the classification result. During training, the cost was derived by categorical cross-entropy function.\cite{Zhang2018} ATan was selected as a surrogate gradient calculation for the SC layer during training.\cite{Eshraghian2022-training} We also used Adam optimizer\cite{Kingma2015} with an initial learning rate between 1$e^{-2}$ and 1$e^{-5}$ and a weight decay between 0 and 1$e^{-3}$ as the model optimizer and ReduceLROnPlateau with a factor of 0.9, patience between 1 and 3, threshold between 1$e^{-4}$ and 1$e^{-6}$, minimum learning rate between 1$e^{-4}$ and 1$e^{-7}$ as the learning scheduler, and selected a set of hyper-parameters based on the best performances. Batch sizes of 32 and 150 epochs were used. All experiments are performed on an Intel Xeon Gold 6226R and an NVIDIA A40.\par

\subsection{Experimental Results}

\begin{figure}[h]
  \centering
  \includegraphics[width=\linewidth]{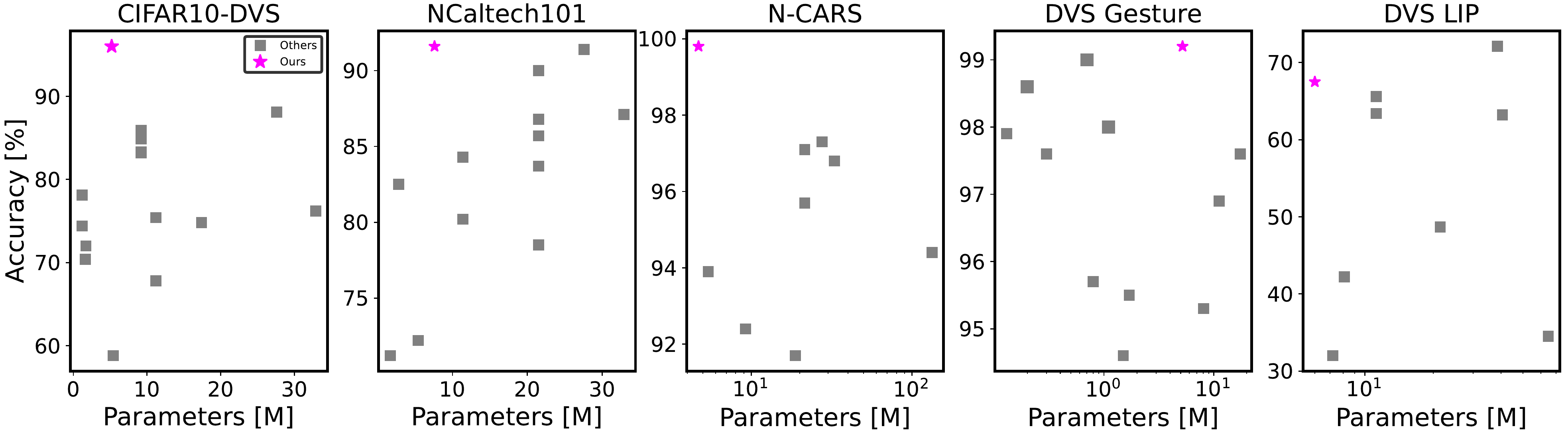}
  \caption{RN-Net on CIFAR10-DVS, N-Caltech101, N-CARS, and DVS Lip compared with existing works in classification accuracy and network size.}
  \label{fig:Comparison}
\end{figure}

\begin{table}[h]
\LARGE
 \caption{Performance and Comparison of RN-Net on CIFAR10-DVS, N-Caltech 101, DVS128 Gesture, and N-CARS datasets.}
  \resizebox{\columnwidth}{!}{%
    \begin{tabular}{llccccc}
        \hline
        Model & Backbone & Params & CIFAR10-DVS & N-Caltech 101 & DVS Gesture & N-CARS\\
              &          &   [M]  &    [\%]     &     [\%]      &   [\%]      & [\%]\\
        \hline
        HATS\cite{Sironi2018}            & Linear SVM &-   & 52.4 & 64.2 &  -   & 90.2\\
        Gao \textit{et al.}\cite{Gao2020}& FLNN       &5.4 & 58.8 & 72.2 &  -   & 93.9\\
        tdBN\cite{Zheng2021}             &ResNet-19   &11.2& 67.8 &  -   & 96.9 & -\\
        SlideGCN\cite{SlideGCN}          &GCN         &  - & 68.0 & 76.1 &  -   & 93.1\\
        LIAF-Net\cite{Wu2022-LIAF}       & -          & 1.6& 70.4 &  -   & 97.6 & -\\\\
        TA-SNN\cite{Yao2021}             & -          & 1.7& 72.0 &  -   & 98.6 & -\\\\
        SEW-ResNet\cite{Fang2021-deep}   &Wide-7B-Net & 1.2& 74.4 &  -   & 97.9 & -\\\\
        PLIF\cite{Fang2021-incorporating}& -          &17.4& 74.8 &  -   & 97.6 & -\\\\
        MVF-Net\cite{MVFNet}             &ResNet-18+34&32.9& 76.2 & 87.1 &  -   & 96.8\\\\
        TCJA-TET-SNN\cite{Zhu2022}       & -          & 9.2& 83.3 & 82.5 & 99.0 & -\\\\
        PSN\cite{PSN}                    &VGG11       & 9.2& 85.9 &  -   &  -   & -\\
        MST\cite{MST}                    &Swin-T      &27.6& 88.1 & 91.4 &  -   & 97.3\\\\
        N-ImageNet\cite{N-ImgNet}        &ResNet-34   &21.5&  -   & 86.8 &  -   & -\\
        She \textit{et al.}\cite{she2022}& -          & 1.7&  -   & 71.2 & 98.0 & -\\
        MatrixLSTM+E2VID\cite{MLSTM}     &ResNet-34   &21.5&  -   & 85.7 &  -   & 95.7\\
        ACE-BET\cite{ACE-BET}            &ResNet-34   &21.5&  -   & 90.0 &  -   & 97.1\\
        Event Clouds\cite{Wang2019}      &PointNet    & 8.1&  -   &  -   & 95.3 & -\\
        Asynet\cite{asynet}              &VGG13       &133 &  -   &  -   &  -   & 94.4\\
        \hline
        \textbf{Ours}                    &\textbf{RN-Net} & \textbf{5.2} & \textbf{96.0} & \textbf{91.6} & \textbf{99.2} & \textbf{99.8}\\
        \hline
    \end{tabular}
  \label{tab:results of 4 datasets}
}
\end{table}

\begin{table*}[h]
\small
 \caption{Comparison of existing models and RN-Net on Lip reading dataset including DVS Lip. The accuracy values are from \cite{Tan2022}}
  \resizebox{\columnwidth}{!}{%
  \begin{tabular}{llcclllll}
    \hline
    Model &   Input & ACC &Params& Backbone & Preprocess& Local encoding & Global encoding\\
    &&[\%]&[M]&&&&&\\
    \hline
    DFTN\cite{Xiao2020}     & Video       & 63.2 & 40.5 & ResNet-18 & - & DFN & BiGRU\\
    Feng et al\cite{Feng2020}     & Video & 63.4 & 11.2 & ResNet-18 & - & - & BiGRU\\
    Martinez et al\cite{Martinez2020} & Video & 65.5 & 11.2 & ResNet-18 & Greyscaling & - & Multi-scale TCN\\
    Event Clouds\cite{Wang2019} & Event & 42.2 & 8.1 & PointNet & Random Sampling & 3D point cloud & 3D point cloud\\
    EV-Gait-3DGraph\cite{Wang2022-event-stream} & Event & 32.0 & 7.2 & - & OctreeGrid Filtering & 3D-Graph & N/A\\
    EV-Gait-IMG\cite{Wang2019-EV-gait}     & Event & 34.5 & 64.6 & - & Event Noise Cancelling & Image-like & N/A\\
    EST\cite{Gehrig2019} & Event & 48.7 & 21.5 & ResNet-34 & Normalized Time Stamp & MLP (Grid) & N/A\\
    MSTP\cite{Tan2022} & Event & 72.1 & 38.5 & ResNet-18 & Multiple Time-Scaling & Voxel Grid & BiGRU\\
    \hline
    \textbf{RN-Net} & \textbf{Event} & \textbf{67.5} & \textbf{6.0} & \textbf{-} & \textbf{-} & \textbf{Reservoir} & \textbf{Reservoir}\\
    \hline
  \end{tabular}%
  }
  \label{tab:results of DVSlip}
\end{table*}

\begin{table}[h]
 \centering
 \caption{Performance comparison of different $\emph{R}_{in}$/$\emph{R}_{f}$ layer configurations for DVS128 Gesture and DVS Lip tasks.}
  \begin{tabular}{cccc}
    \hline
    Local encoding  & Global encoding  & Gesture & Lip  \\
    $\emph{R}_{in}$ &   $\emph{R}_{f}$ & [\%]    & [\%] \\
    \hline
    Time Surface            & Time Surface            & 96.2          & 44.5\\
    Time Surface            & Reservoir Node          & 97.7          & 49.7\\
    Reservoir Node          & Time Surface            & 97.0          & 61.0\\
    \textbf{Reservoir Node} & \textbf{Reservoir Node} & \textbf{99.2} & \textbf{67.5}\\
    Reservoir Node          & Temporal AvgPool        & 96.6          & 62.5\\
    \hline
  \end{tabular}
  \label{tab:Ablation study}
\end{table}

RN-Net shows excellent performance (classification accuracy) on all the datasets. Comparisons of classification accuracy and model size (the number of parameters) with existing networks are shown in Figure~\ref{fig:Comparison} and Tables~\ref{tab:results of 4 datasets} and~\ref{tab:results of DVSlip}. The accuracies of networks listed in Table~\ref{tab:results of DVSlip} were obtained from \cite{Tan2022}, and the parameter size of ResNet variants are obtained from \cite{Leong2020}. RN-Net outperforms all existing networks on CIFAR10-DVS, N-Caltech 101, and N-CARS tasks with much smaller network size, while only using the first 10 percent of the longest clip on DVS128 Gesture. We attribute the improved performance of RN-Net even at a smaller network size to the RN layers' powerful capability to effectively encode temporal information hidden in the event streams. The spatiotemporal features captured at the $\emph{R}_{in}$ and $\emph{R}_{f}$ layers allow more efficient processing by the subsequent DNN blocks and FC blocks, respectively. For DVS Lip, RN-Net achieves top 2 accuracy, only behind MSTP\cite{Tan2022} which is a multi-branch network with different input channels generating frames in different temporal granularity, and employs Voxel Grid as the feature descriptor and using larger BiGRU and ResNet-18 network architectures. By comparison, RN-Net uses a single branch and a much lighter DNN network, without expensive feature description, preprocessing, and recurrent units that can lead to significant hardware and latency overheads.\par

Interestingly, we found that all prior works that showed success for the DVS Lip task are based on DNNs using optimized input representation techniques, and SNN-based works are more successful on the others. This observation seems to suggest that conventional SNNs work well for tasks with temporally coarse and repetitive features, but suffer from tasks where both temporally fine-grained features and global temporal features are critical, such as DVS Lip. We hypothesize that the use of RNs with different time constants to encode temporal features and integrated DNN blocks to capture spatial features in RN-Net allow it to better process spatio-temporal features at both local and global scales. Similar concepts, when implemented in SNNs, may help SNN performance for similar tasks.\par

\subsection{Ablation Study}

The effects of the RN in $\emph{R}_{in}$ and $\emph{R}_{f}$ layers on model performance are investigated in an ablation study by replacing each RN layer with a TS or Temporal average pooling (TAP, only for $\emph{R}_{f}$) layer without modifying the rest of the network. (Table~\ref{tab:Ablation study}) In both tasks, the case with both RN layers shows the best performance, while the case with TS at both places shows the worst performance. However, replacing $\emph{R}_{f}$ with TS degrades the performance more than that of $\emph{R}_{f}$ for the DVS128 Gesture task, while the DVS Lip task shows the opposite tendency. We attribute this difference to different characteristics of the datasets. Data in DVS128 Gesture are repetitive along the clip, which can tolerate less precise input encoding because there are opportunities to capture the lost information at a different time instant. On the other hand, in DVS Lip information at different time instants is unique, which makes input encoding more critical for the network performance.
Interestingly, although the use of RNs achieves the best results for both datasets, the use of TAP in the $\emph{R}_{f}$ layer achieves better results than the use of TS for the DVS Lip task, while the use of TS is better for the DVS128 Gesture task. We speculate that this is caused by the characteristics of TS, which completely discards far history that is critical for the DVS Lip dataset. On the other hand, the repetitive inputs in DVS128 Gesture alleviate this problem of TS implementation, where its capability to encode local temporal information outperforms TAP. As a result, this study illustrates that the RNs’ capability to capture both local and global temporal dynamics allows networks based on them (i.e. RN-Net) to achieve higher accuracy for both tasks.\par

\subsection{Power Estimation}

The power efficiency largely depends on implementation methods of RNs, using either a memristor or a resistor-capacitor (RC) unit.\cite{Du2017,Moon2019,Fang2021} In the following sections, the power of a proposed hardware system running RN-Net for DVS128 Gesture and DVS Lip is representatively estimated based on an existing memristor technology and reported values.\cite{Du2015}\par

\subsubsection{Spike Encoding}

One physical RN device (i.e. a memristor) is assigned to one pixel of the event-based camera. Every time the pixel senses an intensity difference larger than the pre-fixed value, it produces a spike (event), which is input to the dedicated RN. Then, the RN natively encodes temporal information of the spikes following the Equation~\ref{equ:dynamics}.\par

The total spike encoding energy consumed by RNs in the $\emph{R}_{in}$ layer during a clip can be calculated following:\par
\begin{equation}
E_{encoding,in}=\sum_{n=1}^{N} {V_{pulse}}^{2}*G(x_{n},y_{n},p_{n})*t_{pulse}
\end{equation}

where $\emph{E}_{encoding}$ is the total energy consumed in the $\emph{R}_{in}$ layer, N is the total number of input spikes in the video clip, $\emph{V}_{pulse}$ is the (fixed) amplitude of the input spike, $\emph{G}$($\emph{x}_{n}$,$\emph{y}_{n}$,$\emph{p}_{n}$) is the conductance of the RN device assigned to the pixel located at $\emph{x}_{n}$, $\emph{y}_{n}$, with polarity $\emph{p}_{n}$, and $\emph{t}_{pulse}$ is the (fixed) time duration of the spike.\par

Similarly, in $\emph{R}_{f}$, an RN device is assigned to a neuron in the output layer of the convolution blocks. The total spike encoding energy of $\emph{R}_{f}$ during a clip is calculated as:\par

\begin{equation}
E_{encoding,f}=\sum_{i=1}^{N_{f}} {V_{pulse}}^{2}*G(N_{i})*t_{pulse}
\end{equation}
where $\emph{G}$($\emph{N}_{i}$) is the conductance of the RN device assigned with the $\emph{i}$-th neuron in the output layer of convolution blocks along a clip. $\emph{N}_{f}$ is the total number of spikes from the output layer during the state retrieval window.\par
Considering the minimum time interval (1$\mu$s) of events for typical event-based cameras,\cite{Amir2017,Tan2022} we set $\emph{t}_{pulse}$ to 1$\mu$s for both $\emph{R}_{in}$ and $\emph{R}_{f}$. $\emph{V}_{pulse}$ and $\emph{G}_{max}$ are set to 1.5 V and 100 $\mu$S, according to \cite{Du2015}.\par
We simulated the average total energy consumption of $\emph{R}_{in}$ and $\emph{R}_{f}$ for a typical clip in DVS128 Gesture and DVS Lip tasks. Using the total energy and the total number of spikes, the average energy consumption per spike was also derived. The results are shown in Table~\ref{tab:Energy of encoding}.\par

\begin{table}[h]
\centering
\small
 \caption{Total encoding energy and the energy per spike for spike encoding in $\emph{R}_{in}$ and $\emph{R}_{f}$, for a typical DVS128 Gesture and DVS Lip video clip.}
  \begin{tabular}{cccc}
    \hline
    Dataset &   $\emph{R}_{in}$/$\emph{R}_{f}$ & Total Energy & Energy/Spike\\
    &&[$\mu$J]&[pJ]\\
    \hline
    \multirow{2}{*}{DVS128 Gesture} & $\emph{R}_{in}$  & 15.0 & 156.2\\
                                    & $\emph{R}_{f}$   & 1.1  & 123.3\\
    \multirow{2}{*}{DVS Lip}        & $\emph{R}_{in}$  & 0.3  & 22.8 \\
                                    & $\emph{R}_{f}$   & 0.7  & 185.7\\
    \hline
  \end{tabular}%
  \label{tab:Energy of encoding}
\end{table}

\subsubsection{State Retrieval}

The states of RNs in $\emph{R}_{in}$ and $\emph{R}_{f}$ are retrieved for the forward-pass every 30ms and 300ms, respectively. The total retrieval energy of $\emph{R}_{in}$ and $\emph{R}_{f}$ is calculated as:\par

\begin{equation}
E_{retrieval,in}=\sum_{p=0}^{1} \sum_{y=1}^{128} \sum_{x=1}^{128} {V_{read}}^{2}*G(x,y,p)*t_{read}
\end{equation}

\begin{equation}
E_{retrieval,f}=\sum_{i=1}^{D_{7}} {V_{read}}^{2}*G(N_{i})*t_{read}
\end{equation}

where $\emph{E}_{encoding,in}$ and $\emph{E}_{encoding,f}$ are the energy consumption for the state retrieval of $\emph{R}_{in}$ and $\emph{R}_{f}$, respectively, $\emph{V}_{read}$ is the amplitude of the read pulse, $\emph{G(x,y,p)}$ is the conductance of a node located at $\emph{x}_{n}$,$\emph{y}_{n}$, with polarity $\emph{p}_{n}$ in $\emph{R}_{in}$, $\emph{G}$($\emph{N}_{i}$) is the conductance of a node assigned to the $\emph{i}$-th node in $\emph{R}_{f}$, and $\emph{t}_{read}$ is the duration of the read pulse.\par
Different from spike encoding, the read operations do not induce conductance change. For state retrieval, the $\emph{V}_{read}$ amplitude was set to 0.5 V according to \cite{Du2015}, and pulse duration was set as 1$\mu$s for both $\emph{R}_{in}$ and $\emph{R}_{f}$ layers.\par

Based on the energy per read, the number of read retrievals and the number of nodes in both reservoir layers, the average total energy of RN state retrieval during the whole video clip, energy per state retrieval, and energy per node were calculated. The values are presented in Table~\ref{tab:energy of retrieval}.\par

\begin{table}[h]
\centering
\small
 \caption{Total retrieval energy during a DVS128 Gesture and DVS Lip video clip, and the energy per state retrieval and per node, for $\emph{R}_{in}$ and $\emph{R}_{f}$ RN layers.}
  \begin{tabular}{ccccc}
    \hline
    Dataset &   $\emph{R}_{in}$/$\emph{R}_{f}$ & Total & Energy & Energy\\
                                              && Energy & /Retrieval & /Node\\
                                              &&[nJ]&[nJ]&[pJ]\\
    \hline
    \multirow{2}{*}{DVS128 Gesture} & $\emph{R}_{in}$  & 2234.9 & 44.7 & 1.4\\
                                    & $\emph{R}_{f}$   & 24.6   & 4.9  & 9.6\\
    \multirow{2}{*}{DVS Lip}        & $\emph{R}_{in}$  & 178.7  & 3.6  & 0.3\\
                                    & $\emph{R}_{f}$   & 33.4   & 6.7  & 13.0\\
    \hline
  \end{tabular}%
  \label{tab:energy of retrieval}
\end{table}

\subsubsection{Data Conversion}
 The SC layer does not require analog-digital conversion because the SC layer takes digital values from the convolution blocks and outputs a binary value using a threshold value within the digital domain. Analog-digital converters (ADCs) are however needed at the state retrieval process, which converts the retrieved analog values into digital values for the following digital DNN blocks. To be conservative, 8-bit ADCs between the RN output and the digital DNN blocks were considered to provide sufficient precision during conversion.\par

For the DVS128 Gesture dataset, the number of RNs in the $\emph{R}_{in}$ layer is 8192 (2x64x64) and the number of retrievals is 50. Thus, the total number of ADC operations is 409600 during 1.5s running time. It requires at least a 273KSPS (kilo samples per second) ADC. For the $\emph{R}_{f}$ layer, 512 RNs are necessary, and the states are retrieved 5 times during 1.5s running time. It requires 2560 ADC operations during a video, corresponding to a 1.7KSPS ADC. To perform conservation estimate, we chose the power of a commercial 8 bit-ADC (ADC7040 ultra-low power SAR ADC from TI), which consumes 171 $\mu$W for 274.7KSPS of the task.\par

For the DVS Lip dataset, 11552 (2x76x76) RNs are in the $\emph{R}_{in}$ layer and the total number of ADC operations is 577600 (11552x50) during 1.5s running time, which demands at least a 385KSPS ADC. For the $\emph{R}_{f}$ layer, the same ADC used for DVS128 gesture dataset is needed to support the same structure. If we utilize the same commercialized ADC, 171 $\mu$W are additionally consumed for DVS Lip dataset. Furthermore, we expect that the power consumption will become less with less precision ADCs (e.g. 6-bit or 4-bit) which are commonly used in neuromorphic circuits.\par


\subsubsection{Overall System Power Estimation}

The total energy of the RN-Net hardware system can be calculated by summing up the energy used by the $\emph{R}_{in}$/$\emph{R}_{f}$ layers and ADCs calculated above, plus the energy used for the DNN blocks to run a clip for DVS128 Gesture and DVS Lip. To calculate the energy of the DNN blocks, we derived the number of MAC operations for running a video clip in RN-Net, according to Table~\ref{tab:network configuration}. The number of MAC operations during a video clip was calculated based on the number of MAC operations in the convolution and fully-connected layers, multiplied by the number of forward passes (i.e. 50) through these layers during the clip. We use published, conservative TOPS/W value (i.e. 2 TOPS/W from the Google Edge TPU \cite{TPU_edge}) to estimate the energy used by the DNN blocks since these blocks can be implemented in any digital accelerator. By adding all the energy costs, the total energy consumed by RN-Net per video clip, and the average power of RN-Net for DVS128 Gesture and DVS Lip tasks are estimated at 10.3mW and 11.6mW, respectively, and more details are presented in Table~\ref{tab:total energy}.\par

\begin{table}[h]
\centering
\small
 \caption{Total energy and average power of the RN-Net system, along with the total number of MAC operations in the DNN blocks when running DVS128 Gesture and DVS Lip tasks.}
  \begin{tabular}{cccccc}
    \hline
    Dataset & Convolution/ & Fully-    & Whole   & Energy & Power\\
            & Forward-pass & connected & Network & & \\
    &[MOPS]&[MOPS]&[GOPS]&[mJ]&[mW]\\
    \hline
    Gesture & 608.7 & 2.7 & 30.4 & 15.4 & 10.3\\
    Lip        & 686.3 & 2.7 & 34.3 & 17.3 & 11.6\\
    \hline
  \end{tabular}
  \label{tab:total energy}
\end{table}

\section{Conclusion}
\label{sec:formatting}

In this work, we propose a hybrid reservoir/DNN network, Reservoir Nodes-enabled neuromorphic vision sensing Network (RN-Net), for asynchronous event-based vision processing. The use of dynamic reservoir nodes enables efficient spatiotemporal encoding of both local and global features at different hierarchies in real time and eliminates external memory and logic/recurrent units. RN-Net achieves the highest classification accuracy on CIFAR10-DVS, N-Caltech 101, DVS128 Gesture, and N-CARS and one of the highest DVS Lip classification accuracy with a much lighter network structure. The reservoir nodes can be efficiently implemented with STM memristors, taking advantage of internal device physics to perform signal processing. The low hardware and training costs of RN-Net make it an attractive option for event-camera applications. Additionally, new devices, such as optical neural transistors, can potentially both sense optical inputs and process temporal information in one device\cite{neural-opto-transistor}, allowing better power-efficiency in vision processing following approaches discussed here.\par

\medskip

\medskip
\textbf{Acknowledgements} \par 
This work was supported in part by the Semiconductor Research Corporation (SRC) and Defense Advanced Research Projects Agency (DARPA) through the Applications Driving Architectures (ADA) Research Center and in part by the National Science Foundation under Grants CCF-1900675 and ECCS-1915550.

\medskip


\end{document}